\documentclass[conference]{IEEEtran}
\makeatletter
\def\ps@IEEEtitlepagestyle{%
  \def\@oddfoot{\mycopyrightnotice}%
}
\def\mycopyrightnotice{%
  \begin{minipage}{\textwidth}
  \centering \scriptsize
  This work has been accepted for presentation at the 2023 IEEE
International Conference on Robotics and Automation (ICRA), May 29 - June 2, 2023, London, UK. 

arXiv version will be merged with the conference proceeding once available.
  \end{minipage}
}
\IEEEoverridecommandlockouts
\usepackage{cite}
\usepackage{booktabs}
\usepackage{amsmath,amssymb,amsfonts}
\usepackage{algorithmic}
\usepackage{mathtools}
\usepackage{multirow}
\usepackage{graphicx}
\usepackage[ruled,vlined,linesnumbered]{algorithm2e}
\usepackage{textcomp}
\usepackage{xcolor}
\usepackage{float}
\usepackage{bm}
\usepackage{soul}
\usepackage[top=72pt, left=54pt, right=54pt, bottom=57pt]{geometry}
\usepackage{nidanfloat}
\newcommand{\etal}{et al. }

\newcommand{\argmin}{\arg\!\min}
\def\BibTeX{{\rm B\kern-.05em{\sc i\kern-.025em b}\kern-.08em
    T\kern-.1667em\lower.7ex\hbox{E}\kern-.125emX}}
\begin{document}
\pdfoutput=1 

\title{RAMP-Net: A \underline{R}obust \underline{A}daptive \underline{M}PC for Quadrotors via \underline{P}hysics-informed Neural \underline{Net}work}

\author{Sourav Sanyal (\emph{Graduate Student Member, IEEE}) and Kaushik Roy (\emph{Fellow, IEEE})\\
% % \IEEEauthorblockN{Sourav Sanyal}
  \IEEEauthorblockA{Elmore Family School of Electrical and Computer Engineering,
 Purdue University\\
 \{sanyals, kaushik\}@purdue.edu
 }
% % Atlanta, Georgia 30332--0250\\
% % Email: http://www.michaelshell.org/contact.html}
% \and
% \IEEEauthorblockN{Kaushik Roy}
% \IEEEauthorblockA{Twentieth Century Fox\\
% Email: homer@thesimpsons.com}
% \and
% \IEEEauthorblockN{James Kirk\\ and Montgomery Scott}
% \IEEEauthorblockA{Starfleet Academy\\
% San Francisco, California 96678--2391\\
% Telephone: (800) 555--1212\\
}

%Telephone: (800) 555--1212, Fax: (888) 555--1212}


\maketitle
\vspace{-10mm}
\begin{abstract}
Model Predictive Control (MPC) is a state-of-the-art (SOTA) control technique which requires solving hard constrained optimization problems iteratively. For uncertain dynamics, analytical model based robust MPC imposes additional constraints, increasing the hardness of the problem.  The problem exacerbates in performance-critical applications, when more compute is required in lesser time. Data-driven regression methods such as Neural Networks have been proposed in the past to approximate system dynamics. However, such models rely on high volumes of labeled data, in the absence of symbolic analytical priors. This incurs non-trivial training overheads. Physics-informed Neural Networks (PINNs) have gained traction for approximating  non-linear system of ordinary differential equations (ODEs), with reasonable accuracy. In this work, we propose a Robust Adaptive MPC framework via PINNs (RAMP-Net), which uses a neural network trained partly from simple ODEs and partly from data. A physics loss is used to learn simple ODEs representing ideal dynamics. Having access to analytical functions inside the loss function acts as a regularizer, enforcing robust behavior for parametric uncertainties. On the other hand, a regular data loss is used for adapting  to residual disturbances (non-parametric uncertainties), unaccounted during mathematical modelling. Experiments  are performed in a simulated environment for trajectory tracking of a quadrotor. We report 7.8\% to 43.2\% and 8.04\% to 61.5\%  reduction in tracking errors for speeds ranging from 0.5 to 1.75 m/s compared to two SOTA regression based MPC methods.
\end{abstract}


% \fi
\IEEEpeerreviewmaketitle

\section{Introduction}
Model Predictive Control (MPC)\cite{camacho2013model} is an advanced control technique, which involves solving an optimal control problem iteratively, while satisfying a set of constraints. Although traditionally used in oil-refineries and process control \cite{oilrefinery}, with the availability of faster computers, MPC has found widespread popularity in autonomous driving and robotic control \cite{incremona2017mpc}. In order to account for uncertain disturbances (typically encountered in real-life), robust MPC techniques have been proposed which add additional constraints by setting conservative bounds on disturbances during the design phase \cite{tubeMPC}. However, solving the system dynamics accurately in presence of additional constraints within the required time-budget forms a bottleneck, in high-speed applications such as agile drone navigation, even with today's hardware \cite{oravec2017parallel}.

Artificial Intelligence (AI) / Machine learning (ML) based data-driven methods have been put forward, which perform system identification \cite{brunton2016sparse} by fitting kernels obtained through regression methods such as Gaussian processes\cite{GP-MPC} or neural networks\cite{patan2014neural}. These approaches if trained well, relaxes the computational demand during inference, by replacing analytical models with simple kernels \cite{neuro-ising,atr}. However, the purely data-driven AI methods lack explainability\cite{XAI} and may require huge training data, with carefully annotated labels, incurring non-trivial training overheads. Physics-informed neural network (PINN)\cite{raissi2019physics} introduced by  Raissi et. al. approximated system of ordinary differential equations (ODEs) using a neural network. PINNs have emerged as a promising paradigm in the field of numerical optimization \cite{karniadakis2021physics}. The residual of the ODEs are fitted using data to reduce the error using autograd -- an automatic differentiation tool \cite{autograd}, available in  standard neural-network software frameworks. Exploiting this PINN property, the main goal of this work is to perform system identification in context of MPC, via a lightweight neural network with low training overhead.

We propose \emph{RAMP-Net} -- a robust adaptive MPC via PINNs and perform trajectory tracking for a quadrotor in presence of uncertain dynamic disturbances. The PINN is 
trained partly from simple ODEs and partly from data. The ODEs represent the ideal system dynamics of a quadrotor in absence of uncertainties/disturbances.  The data obtained through real-life-like simulated environments (with noises and disturbances) enables the proposed network to adapt to similar disturbances if encountered during inference.  By training from sample sources (called collocation points \cite{raissi2019physics}), whose target labels are obtained through analytical symbolic functions, we are able to infuse system knowledge in the training data. Having access to such analytical functions during training enforces desired system behavior, while also making the model partially interpretable.
% We propose a composite loss function, which results in training convergence while increasing robustness.
% when trained using a quasi-newton optimization approach \cite{quasinewton}. 
The main contributions of this work are as follows:
\begin{itemize}
    \item We formulate the ideal system dynamics of a quadrotor to fit the residual dynamics as a physics loss and use a data loss to capture additional dynamics unaccounted during mathematical modelling (Section \ref{sec3}). 
    % The data loss is responsible for adapting to residual dynamics, on top of the nominal dynamics fitted through the physics loss (Section \ref{sec3}).
    \item We train a PINN using the composite loss (sum of the above mentioned loss functions) to approximate the non-linear dynamics of a quadrotor to propose RAMP-Net -- a robust adaptive MPC via PINNs (Section \ref{sec4}).
     \item We perform trajectory tracking of a Hummingbird quadrotor in the Gazebo simulation environment to obtain $\sim60\%$ lesser tracking error compared to a SOTA regression-based method along with $\sim 11\%$ faster convergence. We report significant reduction in tracking error for various speeds ($0.5 - 12.5$ m/s) w.r.t two SOTA regression based MPC methods \cite{KNODE-MPC, GP-MPC} (Section \ref{res}).
    
\end{itemize}

\vspace{-1mm}
\section{Related Work}
We consider trajectory tracking in the face of uncertain dynamic disturbances. Research endeavours in the past to achieve this are briefly discussed.\\

\noindent{\textbf{Deep Reinforcement Learning based Neural methods:}} Deep Reinforcement Learning (RL) \cite{arulkumaran2017deep, ICRA5} approaches assume the underlying control problem to be a Markov Decision Process (MDP), and uses functional approximation to learn the optimal policy to perform sequential decision making under uncertainty. The authors of \cite{RARL} have combined RL with adversarial learning \cite{goodfellow2014pouget}. The robust optimization problem is addressed using an \emph{actor-critic} setup, where an agent (actor) learns a policy to control the system and another agent (critic) learns a separate policy to destabilize the system. The work in \cite{pan2019risk} extends \cite{RARL} using an ensemble of Deep Q-Networks \cite{DQN-Nature} with the actor being risk-aware and the critic being risk-seeking. Neural-MPC \cite{neuralMPC} uses Deep RL frameworks within an MPC pipeline, and High-MPC \cite{highMPC} exploits RL to learn high level policies from low-level MPC controllers. However, many RL approaches suffer from the sample-inefficiency with lots of training cycles and steady convergence is still a challenge in complex scenarios or in rapidly evolving environments \cite{dai2018sbeed}. Moreover, absence of analytical/symbolic priors results in lack of explainibility \cite{XAI}, making these methods tractable only for simple setups \cite{safe}. 
%In this work, we do not conisder RL based methods subsequently. 
\\

\noindent{\textbf{Data agnostic model-based  Analytical/Symbolic methods:}}
Tube-MPC \cite{tubeMPC} is a typical robust MPC method, which uses a nominal dynamics model and sets conservative bounds of disturbances (called \emph{tube}) on the state variables, to obtain robust behaviour. The conservative uncertainty/disturbance states guarantee stability in the worst-case scenario, however at the expense of increased ``hardness", as more constraint satisfactions are required. This problem exacerbates in performance-critical high-speed applications having strict time-budgets. To reduce the conservatism of robust controllers, adaptive MPC techniques \cite{adaptiveMPC1, adaptiveMPC2} consider parametric uncertainties over state variables. Such techniques either use functional analysis methods to guarantee closed-loop stability or
adapts the controller parameters to mimic a reference model. However, such methods are limited to tackle only parametric uncertainties and tend to overfit to the analytical reference models, a phenomenon known as \emph{model drift}. Hence, model-based adaptive MPC does not guarantee optimal convergence to true parameters.\\

\vspace{-1mm}
\noindent{\textbf{Regression based system identification methods for MPC:} }
We are interested in data-driven control methods in the context of MPC. To achieve this, we consider system identification\cite{brunton2016sparse}, where the analytical model is improved using data-driven regression methods. This has recently inspired researchers to integrate machine learning and MPC \cite{qraitem2020bridging}. To solve this, \cite{GP-MPC} uses Bayesian tools such as Gaussian processes to perform regression. On the other hand, \cite{KNODE-MPC} recently employed a technique called Neural-ODE \cite{NODE} to correct modelled dynamics using neural networks. Our proposed method is along similar direction, where we use PINNs to formulate a composite loss function, with the aim to design a robust adaptive MPC framework, which we now present.
\vspace{-0.3mm}
\section{Proposed Approach}
\label{sec3}
\subsection{\textbf{Dynamical System formulation}}
Let us consider a dynamical system of the form
\begin{equation}
    \dot{\bm{x}}(t) = f(\bm{x}(t),\bm{u}(t))
\end{equation}
on the time interval $\mathbb{T}$ $\in$ $\mathbb{R}$, where $\bm{x}:\mathbb{T} \mapsto \mathcal{X} \in \mathbb{R}^{n}$ represent the state variables and $\bm{u}:\mathbb{T} \mapsto \mathcal{U} \in \mathbb{R}^{m}$ represent the  control variables. In this study, we consider a quadrotor, with $n = 13$ states and $m = 4$ control variables. At $t = 0$, if $x(t) = x[0]$, then we are interested in solving an initial-value problem (IVP) over time-interval $\mathbb{T}$. According to \cite{lip}, if $f$ is Lipschitz-continuous with respect to the state, then the IVP has a unique solution for each $\bm{u}$ $\in$ $L^{\infty}(\mathbb{T}, \mathcal{U})$. From Eqn. (1), we can write

\begin{subequations}
\begin{align}
    \int_{t=k}^{t=k+1}d\bm{x}(t) = \int_{t=0}^{t=T}f(\bm{x}(t),\bm{u}(t))dt \;\; \; \; \; \;\; \; \; \;\\[1pt]
    \implies \bm{x}[k+1] = \bm{x}[k] + \int_{t=0}^{t=T}f(\bm{x}(t),\bm{u}(t))dt \; \; \; \; \;  \\
    \implies  \bm{x}[k+i+1] = \phi(T, \bm{x}[k+i], \bm{u}[k+i]) \; \; \; \; \; \;\; \;\\[1pt]
    \forall i = \{0,1,...,N-1, N\} \;\; \; \; \; \;\; \; \; \; \;\; \; \; \; \;\; \; \; \;
\end{align}
\end{subequations}

\begin{figure}
  \vspace{-0.7cm}
    \begin{center}
        \includegraphics[ width = 0.35\textwidth]{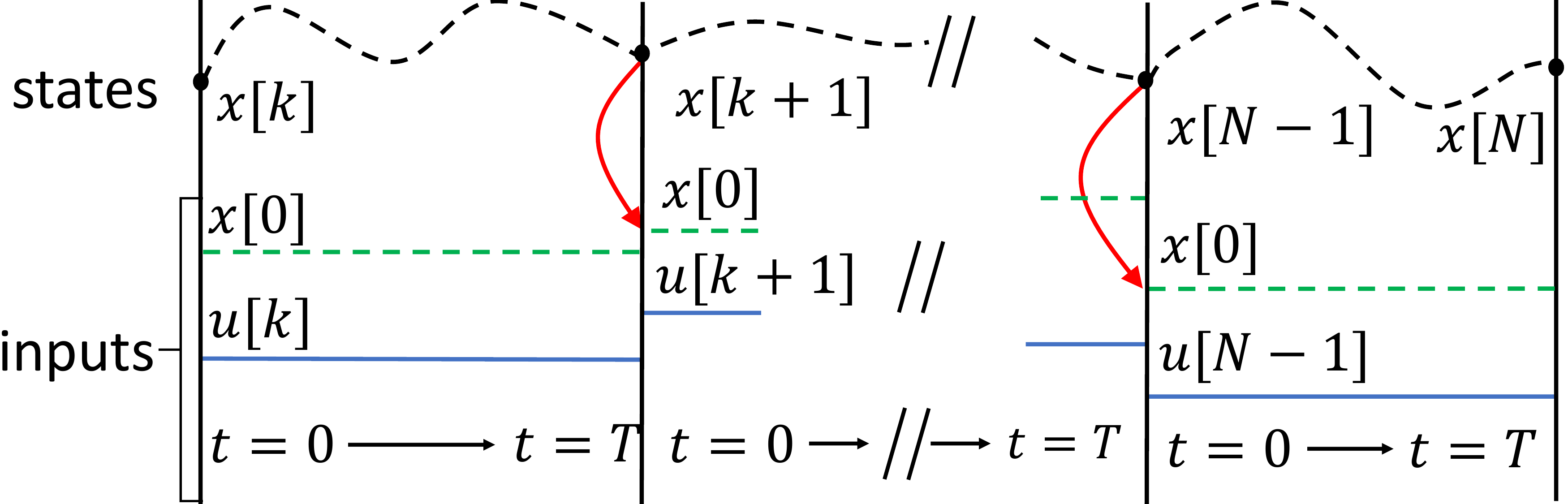}
  \vspace{-0.25cm}
    \end{center}
     \vspace{-0.3cm}
\caption{Moving Horizon Illustration. Best viewed in color.}
 \vspace{-0.51cm}
  \label{MHE}
\end{figure}

Figure \ref{MHE} illustrates the Multiple-Shooting Moving Horizon scheme \cite{biegler2010nonlinear} of MPC, which we adopt. $T$ represents the time-interval. The final state $\bm{x}[k+i]$ of the previous interval is fed as the initial state $\bm{x}[0]$ of the next interval to obtain $\bm{x}[k+i+1]$ by solving an IVP for each interval $T$. This enables a time-discretized implementation of a digital controller. We make a zero-hold assumption for the control variable, i.e,  $\dot{\bm{u}}[k] \equiv 0$, meaning the control signals are constant and discrete  within a moving horizon interval (from $t =0$ to $t= T$). $\phi$ represents a regressor (model-based or data-based or both) to be fitted. 

\subsection{\textbf{MPC problem formulation}}
Given a reference trajectory $\bm{x}^{ref}_k$, and a control system as described in Eqn. (2), we want the state vector ($\bm{x}_k$) of a quadrotor to follow $\bm{x}^{ref}_k$ as closely as possible. We use a quadratic cost $J$ $\in$ $\mathbb{R}$, defined as follows:
\begin{equation}
    J(\bm{x}_k, \bm{u}_k) = \sum_{i=k}^{k+T-1} (||\bm{x}^{ref}_i-\bm{x}^{pred}_i||^2_{\bm{Q}} + ||\bm{u}_i||^2_{\bm{R}})
\end{equation}
where $||\bm{x}||_{\bm{Q}} = \sqrt{\bm{x}^T\bm{Qx}} : \mathbb{R}^n \mapsto \mathbb{R}$ , $||\bm{u}||_{\bm{R}} = \sqrt{\bm{u}^T\bm{Ru}}: \mathbb{R}^m \mapsto \mathbb{R}$ are  weighted semi-norms $\forall$ $\bm{Q}$ $\in$ $\mathbb{R}^{n\times n}$, $\bm{R}$ $\in$ $\mathbb{R}^{m\times m}$ being positive semi-definite. 
The MPC problem then involves iteratively solving the following in real-time:
% \begin{equation}
\begin{subequations}
\begin{align}
    \argmin_{\bm{u}_k} J(\bm{x}_k, \bm{u}_k) \;\; \; \; \; \;\; \; \; \;\\[1pt]
    \text{s.t.} \;\; \; \; \; \bm{x}[k+1] = \phi(T, \bm{x}[k], \bm{u}[k]) \\[1pt]
    \forall \; \;  \bm{x}[k] \in \mathcal{X}, \; \; \bm{u}[k] \in \mathcal{U} \\[1pt]
    \forall \; \;  k \in \{0,1,....,N,N-1\}
    \end{align}
\end{subequations}

 By abuse of notation, $ \bm{x}_k = \bm{x}[k]$, and $\bm{u}_k =\bm{u}[k]$. 
For each interval, the optimal control signals $\bm{u}_k$ are obtained by solving Eqn. (4). Traditionally, model-based MPC employs non-linear programming methods such as interior-point-optimization (IP-OPT) \cite{wright1999numerical} and uses numerical integration schemes such as Runge-Kutta methods. For high speed applications, evaluating $\phi$ can be challenging, specially if additional constraints are imposed for robust tracking, by setting conservative bounds. For complex environments, $\phi$ can be hard to model accurately. In this work, we use a PINN to numerically evaluate $\phi$, enabling rapid system identification for a robust adaptive MPC. We now give a brief background on PINNs and explain how it can be modified to perform system identification of a quadrotor subjected to uncertain dynamic disturbances.

\subsection{\textbf{Physics-informed Neural Networks}}
% \vspace{-2mm}
Physics-informed Neural Networks (PINNs) \cite{raissi2019physics} was introduced in 2019 by Raissi \etal. It solved differential equations by adding the differential equation to the loss function itself, as a residual. If we consider Eqn. (1), we can formulate a residual physics loss as follows:
\begin{equation}
    \mathcal{L}_p =MSE(\dot{\bm{x}}(t),f(\bm{x}(t),\bm{u}(t)))
\end{equation}

$MSE$ stands for the mean square error. The original paper only considered equations with state variables $\bm{x}$. However to solve an MPC, we require addition of control variables $\bm{u}$.  In this work, we add provision for using the control variable $\bm{u}$, and a time variable $t$, to be fed to the neural network, as separate signals, similar to \cite{PINC}. Substituting the continuous variables, with $\phi$ from Eqn. (4), in this work, we arrive at the corresponding physics loss, as shown:
\begin{subequations}
\begin{align}
    \mathcal{L}_p= MSE(\dot{\phi}(T, \bm{x}_k, \bm{u}_k), f(\bm{x}_k, \bm{u}_k)) \; \; \; \;\; \; \; \; \; \; \; \; \;\;  \\
\implies \mathcal{L}_p=MSE(\dot{\phi_T}(\bm{x}_k, \bm{u}_k ; \theta), f(\bm{x}_k, \bm{u}_k)) \; \; \; \; \; \; \;\; \; \; \; \; \\
\implies \mathcal{L}_p = \frac{1}{|\mathcal{P}|}\sum_{k=1}^{|\mathcal{P}|}||\dot{\phi_T}(\bm{x}_k, \bm{u}_k ; \theta)- f(\bm{x}_k, \bm{u}_k)||^2  \; \; \; \;  \\
\forall \; \{\bm{x}_k, \bm{u}_k\} \in \mathcal{P} \; \; \;\; \; \; \; \; \; \;\; \; \; \; \;  \;\; \; \; \; \; \;\; \; \; \; 
\end{align}
\end{subequations}

  \begin{figure}
  \vspace{-0.5cm}
    \begin{center}
        \includegraphics[ width = 0.35\textwidth]{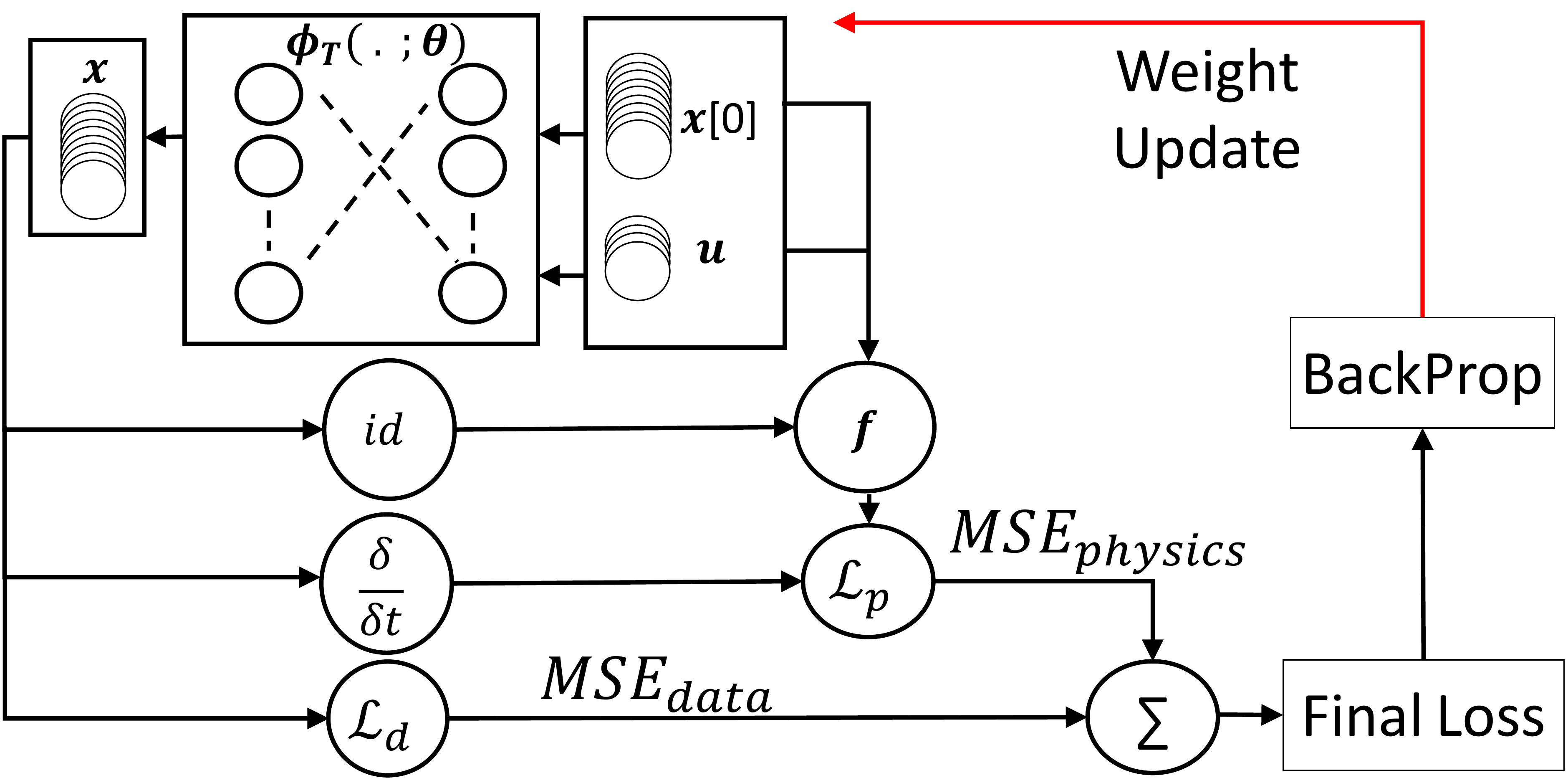}
%   \vspace{-0.5cm}
    \end{center}
     \vspace{-0.3cm}
\caption{PINN Loss = Physics Loss + Data Loss. id implies identity operation.}
 \vspace{-0.51cm}
  \label{loss}
\end{figure}
We use a neural network parametrized by $\theta$ to learn $\phi_T$ and treat $T$ as the network parameter,  dropping it as a function argument. Using autograd\cite{autograd} -- an automatic differentiation tool readily available in standard neural network packages, the numerical derivative $\dot{\phi_T}$ can be easily evaluated in one forward propagation. $\{\bm{x}_k, \bm{u}_k\} \in \mathcal{P}$ are called \emph{collocation points}\cite{raissi2019physics}, where $\mathcal{P}$ is the physics dataset. In the physics loss (Eqn. 6), instead of a labelled target, the symbolic prior applied on the collocation points ($f(\bm{x},\bm{u})$) is used, imposing a constraint on the data loss (Eqn. 7). This constrains/regularizes the neural network to obey the dynamics of a quadrotor modelled in $f$.

Figure \ref{loss} illustrates the implemented PINN loss evaluation.  A simple Multilayer Perceptron (MLP) is trained. The control variable $\bm{u}$ along with the initial values $\bm{x}[0]$ (last measurement from the previous horizon) are fed as inputs. The PINN MLP output is passed through $f$ (the ideal dynamics) and $\frac{\delta}{\delta t}$ 
 (the \emph{autograd} function) to obtain the physics loss. Furthermore, we collect recorded observations to prepare a dataset $\mathcal{D}$ which to fit a data loss as follows:
 \begin{subequations}
 \begin{align}
     \mathcal{L}_d = MSE(\phi_T(\bm{x}_i, \bm{u}_i;\theta), \bm{y}_i) \; \;  \; \; \; \; \; \;  \; \; \; \; \\
     \implies \mathcal{L}_d = \frac{1}{|\mathcal{D}|}\sum_{i=1}^{|\mathcal{D}|}||\phi_T(\bm{x}_i, \bm{u}_i;\theta) - \bm{y}_i)||^2 \; \; \; \;\; \; \\
     \forall \; \{ (\bm{x}_i, \bm{u}_i), \bm{y}_i\} \in \mathcal{D} \; \; \; \;\; \; \; \; \; \; \; \;\; \; \; \;
     \end{align}
 \end{subequations}
   The dataset $\mathcal{D}$ consists of uncertain dynamic disturbances, obtained through noise injections in a simulated physics engine. The data samples $\{\bm{x}_i, \bm{u}_i\} \in \mathcal{D}$ represent the state and control input measurements obtained through odometry, and $\{\bm{y}_i \}$, the corresponding groundtruth label of the quadrotor state derivatives.
  The physics loss and data loss are added together to obtain the final composite PINN loss ($\mathcal{L}_p + \mathcal{L}_d$), which is backpropagated to perform the weight updates. Later, we show quantitatively the relative impact of varying $|\mathcal{P}|$ and $|\mathcal{D}|$ (Section \ref{dp}).

\section{System Design}
\label{sec4}

\begin{figure}
  \vspace{-0.7cm}
    \begin{center}
        \includegraphics[ width = 0.43\textwidth]{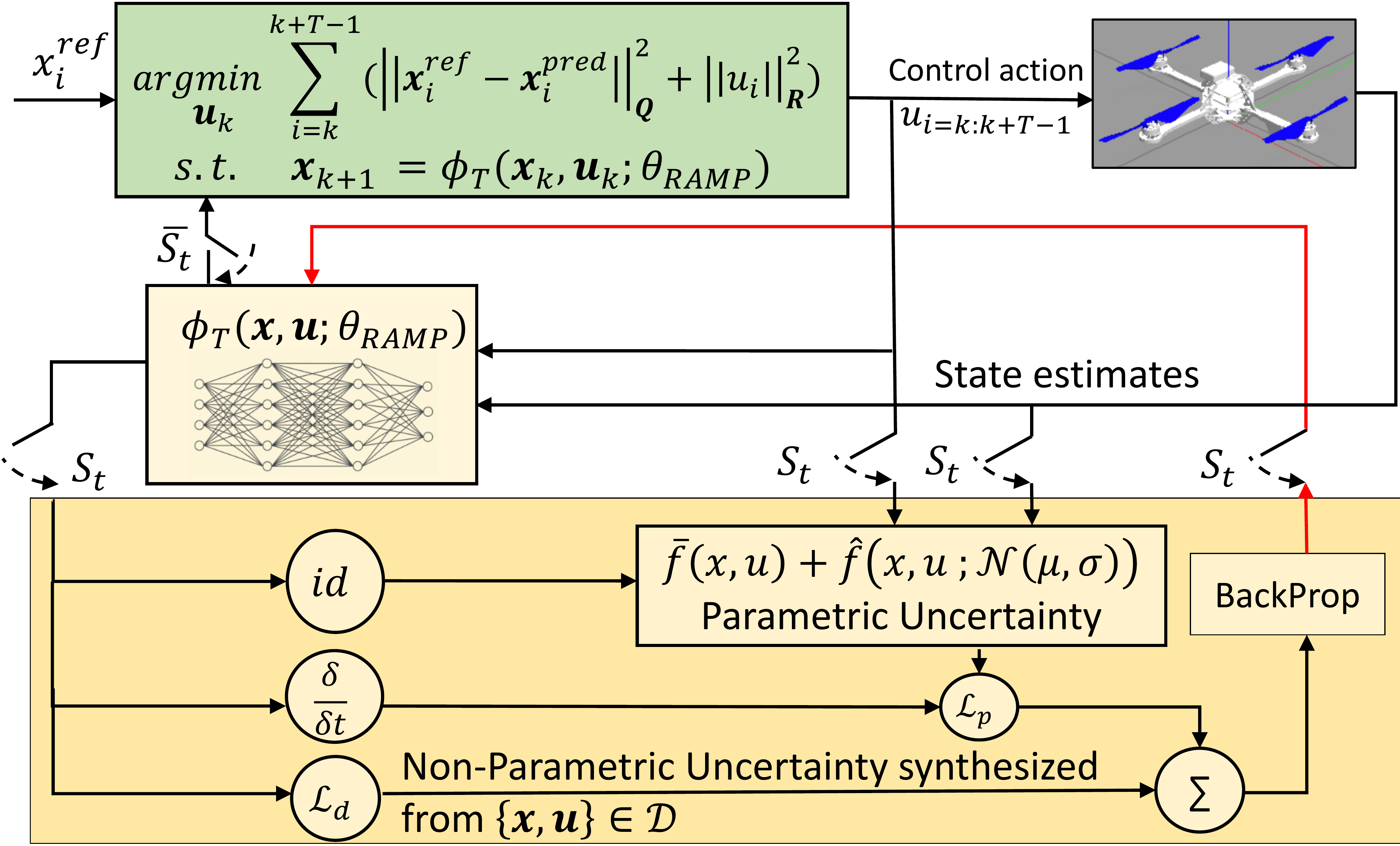}
  \vspace{-0.3cm}
    \end{center}
     \vspace{-0.2cm}
    \caption{RAMP-Net Architecture. Best viewed in color.}
 \vspace{-0.51cm}
  \label{RAMP}
\end{figure}

% We propose \emph{RAMP-Net} -- a Robust Adaptive Model Predictive Control Framework via physics-informed Neural Network.
% As discussed, RAMP-Net is enabled by two contrasting loss functions, each having  distinct roles in achieving robustness.
Figure \ref{RAMP} illustrates the logical view of the proposed RAMP-Net architecture. 
% The bottom shaded part in yellow represents the training block.
% This can be seen as the cloud, where offline learning happens.
A switch $S_t$ is used to toggle between the training ($S_t =$ \textbf{ON}) and inference ($\bar{S_t} =$ \textbf{ON}). We summarize the quadrotor nominal dynamics model similar to \cite{snap,Kamel2017} and subsequently describe how we achieve robust behavior to tackle the issue of uncertain dynamics in context of MPC, when $S_t$ is set to \textbf{ON}.

\subsection{\textbf{Quadrotor Nominal Dynamics}}
Consider a six degrees-of-freedom quadrotor with mass $m$ and diagonal moment of inertia $\bm{J} = \text{diag}(J_x, J_y, J_z)$. We define   $\bm{x} = [\bm{p}, \bm{q}, \bm{v}, \bm{\omega}_B]$ $\forall$ $\bm{x}$ $\in$ $\mathbb{R}^{13}$ as the quadrotor state variables. $\bm{p} = [x,y,z]^T$ $\forall$ $\bm{p}$ $\in$ $\mathbb{R}^3$ are the quadrotor position coordinates in the world frame.  We use the unit quaternions $\bm{q}$ $\in$ $\mathbb{R}^4 = [q_w, q_x, q_y, q_z]^T$ such that $||\bm{q}||=1$ to represent the quadrotor attitude, also in the world frame. $\bm{v}$ $\in$ $\mathbb{R}^3$ are the linear velocities, i.e. $\bm{v} = \dot{\bm{p}}$ in the world frame, and $\bm{\omega}_B$ $\in$ $\mathbb{R}^3$ denotes the angular velocities along XYZ axes in the body frame. We model the quadrotor thrusts $T_i$ $\forall$ $i$ $\in$ $\{0,1,2,3\}$ as the control input signals $\bm{u}$ $\in$ $\mathbb{R}^4$. The quadrotor nominal dynamics is modelled as follows:
\begin{subequations}
\begin{align}
    \dot{\bm{x}} = \begin{bmatrix}
                    \dot{\bm{p}}\\
                    \dot{\bm{q}}\\
                    \dot{\bm{v}}\\
                    \dot{\bm{\omega}_B}
                    \end{bmatrix} 
    = \Bar{f}(\bm{x},\bm{u}) = \begin{bmatrix}
                             \bm{v}\\
                             \bm{q}.\begin{bmatrix}
                                    0\\
                                    \bm{\omega}_B/2
                                    \end{bmatrix}\\
                            \frac{1}{m}\bm{q}\odot \bm{T}_B + \bm{g}\\
                            \bm{J}^{-1}(\bm{\tau}_B - \bm{\omega}_B \times \bm{J}\bm{\omega}_B)
                               \end{bmatrix}\\[4pt]
   \bm{g} = \begin{bmatrix}
            0\\
            0\\
            -9.8 
            \end{bmatrix} \text{, }
   \bm{T}_B = \begin{bmatrix}
            0\\
            0\\
            \sum T_i
             \end{bmatrix} \; \;  \; \; \; \; \; \;  \; \; \; \;\\[4pt]
             \bm{\tau}_B = \begin{bmatrix}
                L (-T_0 -T_1 + T_2 +T_3)\\
                L(-T_0 + T_1 + T_2 - T_3)\\
                k_{drag}(-T_0+T_1 - T_2 + T_3)
                 \end{bmatrix} \; \;  \; \; \; \; \; \;  \; \; \; \;
\end{align}
\end{subequations}
where $\bm{T}_B$ is the collective thrust acting upward, $\bm{\tau}_B$ is the body torque, $k_{drag}$ is the drag constant and $L$ is the quadrotor arm length in $\times$ configuration. $\bm{q}\odot \bm{T}_B$ indicates the rotation of vector body torque by the quadrotor attitude, i.e., $\bm{q}\odot\bm{T}_B = \bm{q}\bm{T}_B\Bar{\bm{q}}$, where $\bar{\bm{q}}$ is the conjugate quaternion.

\vspace{-1.2mm}
\subsection{R\textbf{obustness through infusing Parametric Uncertainty}}
 We modify the system dynamics governed by function $f$ (see Eqn. (1)) as follows:
\begin{equation}
    f(\bm{x},\bm{u}) = \Bar{f}(\bm{x},\bm{u}) + \hat{f}(\bm{x},\bm{u})
\end{equation}
where $\Bar{f}(\bm{x},\bm{u})$ represents the nominal dynamics in Eqn. (8) and $\hat{f}(\bm{x},\bm{u})$ represents the additive parametric uncertainty representing deviations from the nominal quadrotor state variables.
Such deviations can be sampled from standard distributions, such as a normal distribution parameterized by  $\mathcal{N}(\mu, \sigma)$,  $\mu$ and $\sigma$ being the mean and standard deviations respectively (see Section \ref{subsec:method}). This additive term is included in the symbolic prior, influencing the physics loss $\mathcal{L}_p$ as:
\begin{subequations}
\begin{align}
    \mathcal{L}_p = \frac{1}{|\mathcal{P}|}\sum_{k=1}^{|\mathcal{P}|}||\dot{\phi_T}(\bm{x}_k, \bm{u}_k ; \theta_{RAMP})- (\bar{f} + \hat{f})||^2 \\
    \forall \; \bar{f} =  \bar{f}(\bm{x}_k, \bm{u}_k) \; \;  \; \; \; \; \; \;  \; \; \; \;\\
    \forall \;  \hat{f} =  \hat{f}(\bm{x}_k, \bm{u}_k ; \mathcal{N}(\mu, \sigma)) \; \;  \; \; \; \; \; \;  \; \; \; \;
\end{align}
\end{subequations}
% This induces a robust behavior by imposing probabilistic bounds over the state variables.
\vspace{-2mm}
\subsection{\textbf{Adaptation  to Residual Non-Parametric Uncertainty}}
 In addition to analytical modelling (which includes the physics loss), we utilize  data-driven methods to obtain more accurate dynamics. External environment conditions such as winds, disturbances and frictional effects on rotor inertia cannot be parameterized using quadrotor states. We add a non-parametric term $\Tilde{f}$ in Eqn. (9) as follows:
 \begin{equation}
    f_{true}(\bm{x},\bm{u}) = \Bar{f}(\bm{x},\bm{u}) + \hat{f}(\bm{x},\bm{u}) + \Tilde{f}
\end{equation}
where $\Tilde{f}:\mathbb{T} \mapsto \mathcal{X} \in \mathbb{R}^{n}$. We fly the quadrotor in a simulated environment which injects various noises and disturbances. Specifically, we inject zero-mean Gaussian noises with $1$ standard deviation on the rotor thrusts, a 2nd order polynomial aerodynamic drag effect and add asymmetric motor voltage noises in the simulated environment while preparing the dataset $\mathcal{D}$. From Eqn. (4b), using shorthand, we have
\begin{subequations}
\begin{align}
    \bm{x}_{k+1} = \phi_T(\bm{x}_k, \bm{u}_k ; \theta_{RAMP})  \; \;  \; \; \; \; \; \;\\
    \implies  \bm{x}_{k+1} = \bm{x}_{k} +  \int_{t=0}^{t=T}f_{true}(\bm{x},\bm{u})dt \\
    = \bm{x}_{k} +  \int_{t=0}^{t=T}(\bar{f}+\hat{f})(\bm{x},\bm{u})dt +  \Tilde{f}
\end{align}
\end{subequations}
$\Tilde{f}$ is hence synthesized from the dataset $\mathcal{D}$ affecting the data loss $\mathcal{L}_d$, which is rewritten as:
\begin{equation}
    \mathcal{L}_d = \frac{1}{|\mathcal{D}|}\sum_{k=1}^{|\mathcal{D}|}||\bm{x}_{k+1} - \bm{x}_{k} - \int_{t=0}^{t=T}(\bar{f}+\hat{f})(\bm{x}_k,\bm{u}_k)dt ||^2
\end{equation}
 $\mathcal{D}=[((\bm{x}_1, \bm{u}_1), \bm{y}_1), ((\bm{x}_2, \bm{u}_2), \bm{y}_2), ..., ((\bm{x}_{|\mathcal{D}|}, \bm{u}_{|\mathcal{D}|}), \bm{y}_{|\mathcal{D}|}) ]^T$ logged at times $[t_1, t_2, ...., t_{|\mathcal{D}|}]$, where $\bm{y}_k$ is the integrand in Eqn. (13) $\forall$  $k$ $\in$ $\{1,2,....,|\mathcal{D}|\}$. \\
 
  We add the two losses $\mathcal{L}_p$ and $\mathcal{L}_d$ to train our PINN $\theta_{RAMP}$ in order to identify our perceived dynamics in inference mode (when $\bar{S_t}$ is set to \textbf{ON}) as:
 \begin{equation}
     \bm{x}_{k+1} = \phi_T(\bm{x}_k, \bm{u}_k; \theta_{RAMP}).
 \end{equation}
 
\section{Results}
%  \vspace{-1.5mm}
 \label{res}
 \subsection{\textbf{Methodology}}
 \label{subsec:method}
 \vspace{-1.2mm}
We implemented the PINN using Tensorflow \cite{tensorflow2015-whitepaper}, following the approach in \cite{jonas}.
The symbolic nominal dynamics of the quadrotor were implemented in ACADOS\cite{Houska2011a} which wraps around the non-linear optimization toolkit CasAdi \cite{Andersson2019}.
We used a $4$ layer MLP with $128$ neurons in each layer as our \hspace{0.45mm} PINN architecture, and \hspace{0.45mm} trained for $2000$ epochs using early stopping \cite{Prechelt97earlystopping} with a patience of $500$ epochs. We used a learning rate of $1$. For parametric uncertainty, we used a zero mean  normal distribution with unit standard deviation ($\mathcal{N}(\bm{0}, \bm{\Sigma)}$), where $\bm{\Sigma}$ is a constant unit diagonal co-variance matrix. We used the entire dataset as a single batch using the memory-efficient quasi-newton L-BFGS optimizer \cite{quasinewton} following \cite{raissi2019physics, karniadakis2021physics}. We used $5k$ sample points as $|\mathcal{P}| + |\mathcal{D}|$. The weighted coefficients for $\mathcal{L}_p$ and $\mathcal{L}_d$ were set to unity. The PINN based RAMP-Net framework was tested on closed-loop trajectory tracking experiments in the Gazebo \cite{gazebo} environment using the AscTec Hummingbird quadrotor model from the RotorS framework \cite{rotorS}. Table \ref{table1} presents the implementation details.
\vspace{-1mm}
% An autopilot implemented as a low-level PD
% controller\cite{Faessler17ral} was used to spawn the quadrotor in the Gazebo world. 

\begin{table}[h]
\scriptsize
\vspace{-2.2mm}
\centering
\caption{RAMP-Net Implementation Details}
% \resizebox{\columnwidth}{!}{
\begin{tabular}{c|c}
\hline
\textbf{Quadrotor Property}            & \textbf{Value}                                \\ \hline
Mass (m)                               & ($0.68 + 4 \times 0.009$) kg                         \\\hline 
Arm Length ($L$)                         & $0.17$ m                                        \\ \hline
Drag Constant ($k_{drag}$)             & $8.06e-05$                                      \\ \hline
Max Rotor Speed                     & $838$ rad/s                                     \\ \hline
($J_x$, $J_y$, $J_z$)                  & ($0.007$, $0.007$, $0.012$) $kg m^2$ \\ \hline \hline
% \textbf{Autopilot Settings}            & \textbf{Value}                                \\ \hline
% Publish Frequency                      & $200$ Hz                                        \\ \hline
% ($Kp_{xy}, Kd_{xy}$)                   & ($10$, $4$)                                       \\ \hline
% ($Kp_z, Kd_z $)                        & ($15$, $6$)                                       \\ \hline
% ($K_{rp}, K_{yaw}$)                    & ($12$, $5$)                                        \\\hline \hline
\textbf{MPC Settings}                  & \textbf{Value}                                \\ \hline
Time Horizon ($T$)                       & $1$ sec                                         \\ \hline
Publish Frequency                      & $500$ Hz                                        \\ \hline
$\bm{Q}[(0,0):(2,2)]$         & $0.5$                                           \\ \hline
$\bm{Q}[(3,3):(6,6)]$  & $0.1$                                           \\ \hline
$\bm{Q}[(7,7):(9,9)]$ & $0.05$                                          \\ \hline
$\bm{Q}[(10,10):(12,12)]$   & $0.01$                                          \\ \hline
$\bm{R}$                               & diag($0.1$,$0.1$,$0.1$,$0.1$)                         \\ \hline
\end{tabular}
% }
\label{table1}
\end{table}

% ($Kp_{xy}, Kd_{xy}$), ($Kp_z, Kd_z$), and ($K_{rp}, K_{yaw})$ corresponds to the P and D gains in the XY and Z planes, and the P gains for roll-pitch and yaw respectively. 
\vspace{-2.2mm}
\subsection{\textbf{Comparitive Schemes}}
 \begin{itemize}
     \item \textbf{KNODE-MPC \cite{KNODE-MPC}}: This work uses Neural ODEs \cite{NODE} to learn the residual dynamics which is added to a nominal dynamics model.
     \item \textbf{{GP-MPC}}: This scheme employs a Gaussian Process to learn the residual dynamics as a posterior probability distribution, given a prior nominal dynamics.
     \item \textbf{{Ideal}}: This considers the nominal dynamics only as the ideal case, but with perfect oracle of the uncertainties applied as the residual dynamics.
     \item \textbf{{Nominal}}: This considers the nominal dynamics only without any data-driven correction. 
     \item \textbf{{PID}}: This assumes an oracle of the uncertainties and considers a perfectly fine-tuned adaptive PID controller.
 \end{itemize}

\subsection{\textbf{Impact of data on training performance}}
\label{dp}
 We evaluate the training performance in terms of 1) tracking error and 2) total training time. We model the tracking error as a similarity distance between the predicted and desired trajectories using the dynamic time warping (DTW) algorithm \cite{dtw} implemented using \cite{tslearn}, using a metric called data-skewness ($|\mathcal{D}|/|\mathcal{P}|$) which is the ratio of number of regular data points w.r.t the number of collocation points. 
 
\begin{figure}
%   \vspace{-0.5cm}
    \begin{center}

        \includegraphics[ width = 0.47\textwidth]{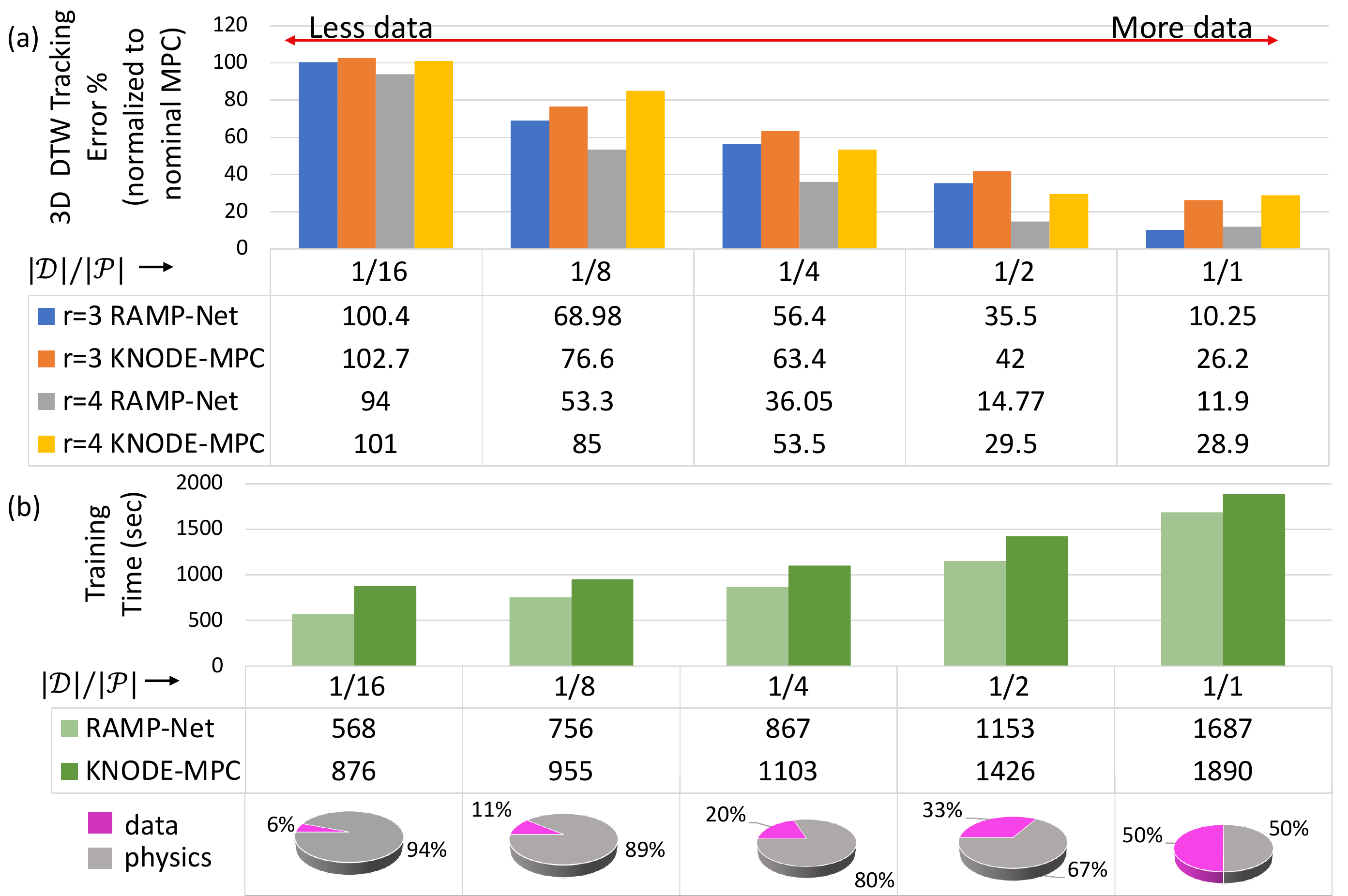}
%   \vspace{-0.5cm}
    \end{center}
     \vspace{-0.5cm}
\caption{Role of data vs physics points on training performance for 2k epochs (a) 3D DTW Tracking Error (b) Training Time  (sec). Best viewed in color.}
 \vspace{-0.51cm}
  \label{d_p}
\end{figure}

 Figure \ref{d_p} illustrates the impact of varying data-skewness. For KNODE-MPC, this translates to the sample complexity as discussed in the paper, where the nominal dynamics is disjoint from the neural network. We swipe from a highly skewed dataset ($1/16$) to zero skew ($1/1$), with intermediate degrees of skewness ($1/8$, $1/4$, $1/2$). We plot the \% DTW errors normalized w.r.t. the nominal baselines for RAMP-Net and KNODE-MPC for circular trajectories of radius $3m$ and $4m$  along with the training times in seconds. The results obtained are for maximum radial velocity of $1 m/s$ at a steady height of $\sim 1m$.  For $|\mathcal{D}|/|\mathcal{P}| = 1/16$, we observe that 3 out of 4 \% DTW errors exceed 100\% over nominal. This indicates that with extreme skewness (very less data), the PINNs do not acquire sufficient expressive power for system identification. With reduction in skewness, we observe: 
 
 1) Errors reduce for both methods, and 2) The error reduction in RAMP-Net is greater than KNODE-MPC with decreasing skewness. For zero skewness, we report $\sim 61\%$ and $\sim 59\%$ lesser errors for circles with radii $3m$ and $4m$ respectively, compared to KNODE-MPC, with $\sim 11 \%$ faster convergence, on the entire training set. Increasing the number of data-points beyond zero-skew (i.e. $|\mathcal{D}|/|\mathcal{P}| > 1$) can overfit to the odometry data and also increases the training overhead. Hence, for subsequent evaluations, we chose a balanced dataset with no skew (i.e. $|\mathcal{D}|/|\mathcal{P}| = 1$).

\subsection{\textbf{Comparison with SOTA Regression based  Methods}}
\begin{figure}[t]
%   \vspace{-0.5cm}
    \begin{center}       \includegraphics[width = 0.4\textwidth]{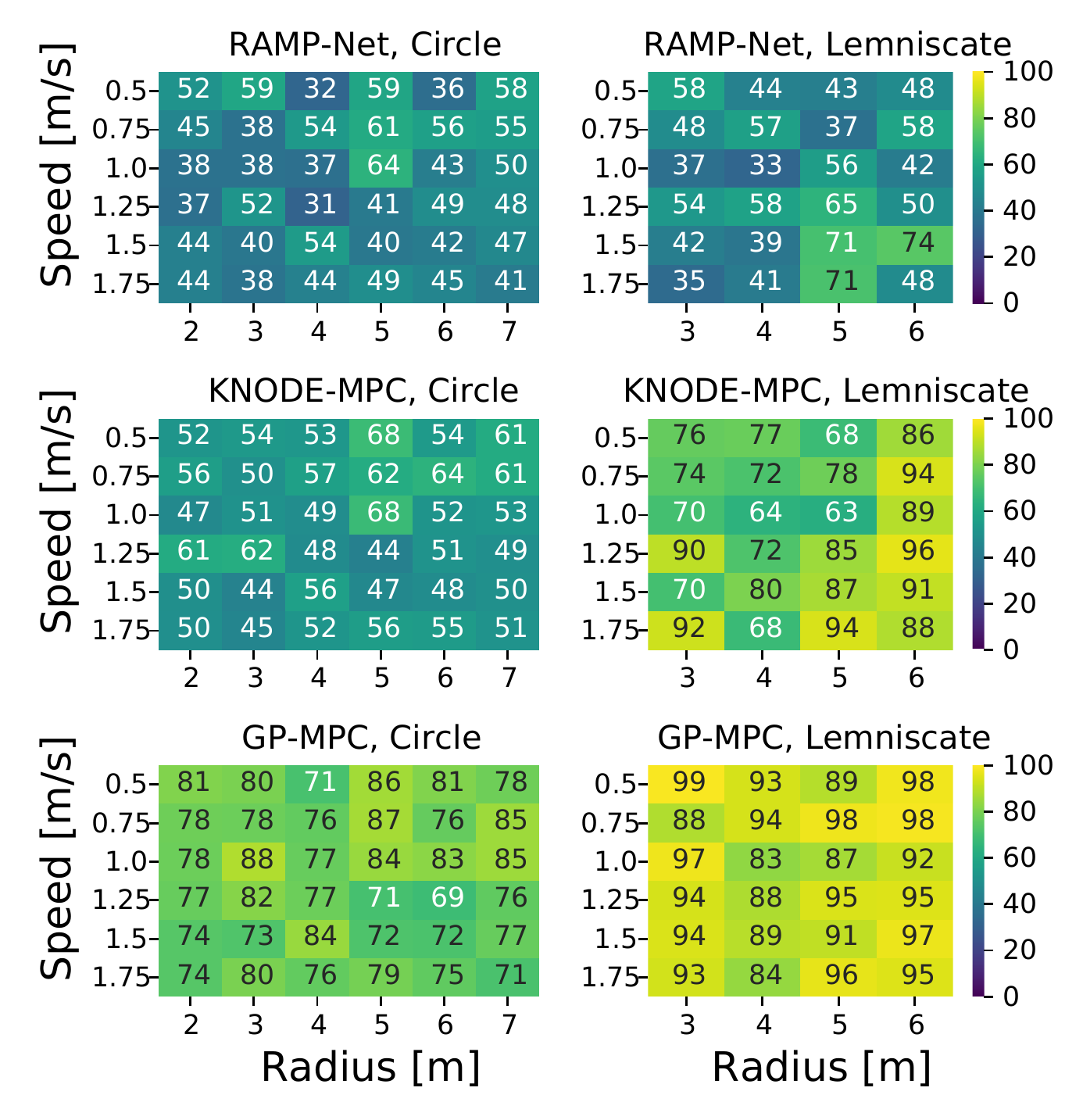}
  \vspace{-0.2cm}
    \end{center}
     \vspace{-0.5cm}
\caption{Heatmap of DTW errors normalized to nominal. Lower is better. \\Best viewed in color.}
 \vspace{-0.6cm}
  \label{pinnloss}
\end{figure}

We compare our work with two other regression based system identification methods for MPC -- KNODE-MPC, and GP-MPC. Figure \ref{pinnloss} presents the \% DTW errors of the three methods for circle and lemniscate trajectories, normalized w.r.t nominal MPC. Though the methods are trained only on circles with radii $3m$ and $4m$, the evaluation is performed for lemniscate trajectories as well, to showcase the generalization capability of data-driven techniques aiding MPC. The horizontal axes in Figure \ref{pinnloss} represent different trajectory radii, while the vertical axes represent the top radial speed. We considered stable hovering at $1m$ for the experiments.

The errors are more in most cases for lemniscate compared to circle trajectory. This is due to the lemniscate shape being more complex than a simple circle, places higher demands on the motor control (more yawing is needed) in presence of environment disturbances. For RAMP-Net, we observe higher variation with increasing speed across different radii ($8.45$, $6.24$, $4.98$ standard deviations  for circle and $11.65$, $10.17$, $4.42$ standard deviations for lemniscate for RAMP-Net, KNODE-MPC and GP-MPC respectively). However, RAMP-Net outperforms both KNODE-MPC and GP-MPC in terms of \% DTW errors normalized to nominal MPC ($46.07 \%$, $53.63 \%$, and $78 \%$ for circle and $47.41\%$, $80.65\%$, and $92.79\%$ for lemniscate for RAMP-Net, KNODE-MPC and GP-MPC respectively, on average). For speeds ranging from $0.5$ m/s to $1.75$ m/s, RAMP-Net outperforms KNODE-MPC and GP-MPC by $7.8 \% - 43.2 \%$, and by $8.04 \% - 61.5 \%$ respectively, on average, in terms of DTW error.

\subsection{\textbf{Robustness Analysis for higher flight speeds}}
\begin{figure}[t]
  \vspace{-0.2cm}
    \begin{center}
        \includegraphics[ width = 0.48\textwidth]{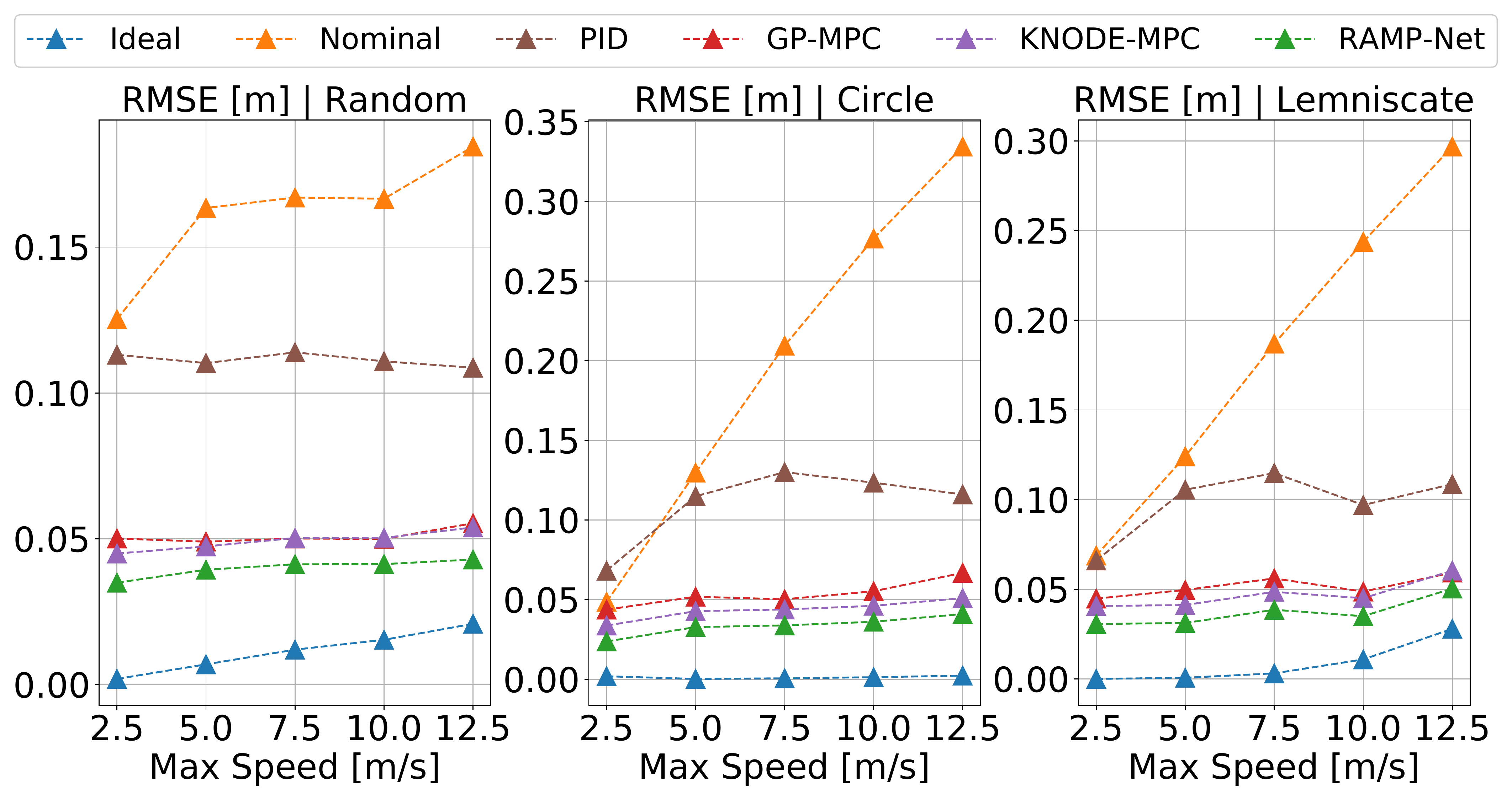}
%   \vspace{-0.5cm}
    \end{center}
     \vspace{-0.4cm}
\caption{RMSE Errors. Lower is better. Best viewed in color.} 
 \vspace{-0.51cm}
  \label{final}
\end{figure}

Figure \ref{final} presents the root-mean-square errors (RMSE) of all the comparative schemes  for radius = $3$m, height = $1$m. Table \ref{tab:my-table} reports the corresponding relative increase in RMSE from ideal, i.e., $|x - ideal|/ideal$, where $x$ is a comparative scheme.  We vary the maximum radial speed as $2.5, 5.0, 7.5, 10.0, 12.5$ m/s. The environment consists of dynamic disturbances along with wind and translational drag effects on the rotors. For the ideal case, the dynamic disturbances are perfectly countered, resulting in the least tracking error. However, with increasing maximum speed, the available time budget reduces, increasing the tracking error. For nominal, there is no corrective measure to counter the disturbances, causing the highest increase. The PID control considers the best tuned gains, however the PID errors are still considerably high ($\sim 27\times$ from ideal). We observe that the errors for GP-MPC, and KNODE-MPC are uncorrelated with maximum radial speed (within $\sim15\times$, and $\sim13\times$ from ideal). RAMP-Net offers the best adaptation, with least tracking error (within $\sim10\times$ from ideal), with sub-linear increase (similar to the GP-MPC and KNODE-MPC), with increasing speed (disturbances), indicating robust tracking.

% \usepackage{graphicx}
\begin{table}[h]
\scriptsize
\centering
\caption{Relative Average RMSE increase normalized to Ideal.\\
Lower is better.}
\label{tab:my-table}
\begin{tabular}{cccccc}
\hline
          & Nominal & PID    & GP-MPC & KNODE-MPC & RAMP-Net \\\hline
Random     & 4.61   & 3.59  & 1.49   & 1.38       & \textbf{1.05}     \\
Circle     & 44.23  & 29.81 & 13.35  & 11.02     & \textbf{8.43}    \\
Lemniscate & 51.45  & 46.67 & 30.01 & 26.97    & \textbf{20.34}  \\\hline
Average & 33.43 &  26.67 & 14.95 &13.12 & \textbf{9.94} \\\hline
\end{tabular}
\vspace{-3mm}
\end{table}

\subsection{\textbf{Latency comparison with standard integrators}}
Table \ref{tab} presents the execution run-times obtained using RAMP-Net and other standard integration methods like Eulers and Runge-Kutta (RK4, RK45) methods, used in MPC without  data-driven regression. We report an order of magnitude lower latency. The times reported for the RAMP-Net PINNs are the wall-clock times observed when running Tensorflow on an NVIDIA GeForce RTX 2080 Ti GPU with 4 cards, 1 GB memory and a clock of 300 MHz. We expect higher speedup with dedicated accelerators and lower-level software routines such as BLAS in C/C++.

% \vspace{-0.25cm}
\vspace{-1mm}

\begin{table}[h]
\scriptsize
\centering
\caption{Execution Time for 1 forward propagation}
\begin{tabular}{ccccc}
\hline
       & RAMP-Net & Euler    & RK4      & RK45    \\ \hline
Mean (sec)  & 4.14e-04 & 5.25e-03 & 2.6e-03  & 9.4e-03 \\
Median (sec) & 2.67e-04 & 5.09e-03 & 2.52e-03 & 5.4e-03 \\ \hline
\end{tabular}
 \vspace{-0.25cm}
\label{tab}
% \vspace{-0.25cm}
\end{table}

\section{Conclusion}
Pure model based robust MPC techniques suffer performance degradation when subjected to uncertain dynamic disturbances (Nominal case in Figure \ref{final}). To that effect, we proposed RAMP-Net -- a robust adaptive MPC framework which uses a neural network that embeds the system (in our case it is a quadrotor) dynamics directly in the neural network loss forming a composite loss function. 
%The physics loss is responsible for robustness against parametric uncertainties by imposing bounds on quadrotor state variables. The data loss is optimized for adaptation to residual non-parametric uncertainties in noisy environments.
Experiments performed on a Hummingbird quadrotor in the Gazebo simulation environment reveal that our proposed method results in $\sim 60\%$ lesser tracking error while training compared to a SOTA regression based method \cite{KNODE-MPC} along with $\sim 11\%$ faster convergence. We report significant reduction in tracking error for various speeds ($0.5 - 12.5$ m/s) compared to two SOTA regression based MPC methods \cite{KNODE-MPC, GP-MPC} and three standard controllers, with faster dynamics integration compared to traditional numerical integration methods. The results establish the effectiveness of incorporating physics-based AI models for solving optimal control problems in noisy settings. This potentially should allow researchers to combine first-principle models with neural networks to better identify real-world dynamical systems.

\section{Acknowledgement}
This work was supported in part by the Center for Brain-inspired Computing (C-BRIC), a DARPA sponsored JUMP center, the Semiconductor Research Corporation (SRC), the National Science Foundation, the DoD Vannevar Bush Fellowship, and IARPA MicroE4AI.

\bibliographystyle{./bibliography/IEEEtran}
\bibliography{./bibliography/main}

\end{document}